\pdfoutput=1
\documentclass[11pt]{article}

\usepackage{ACL2023}
\usepackage{times}
\usepackage{latexsym}
\usepackage[T1]{fontenc}
\usepackage[utf8]{inputenc}
\usepackage{microtype}
\usepackage{inconsolata}

\usepackage{hyperref}
\usepackage{url}
\usepackage{booktabs}
\usepackage{amsfonts}
\usepackage{nicefrac}
\usepackage{xcolor}
\usepackage{color}
\usepackage{amsmath}
\usepackage{adjustbox}
\usepackage{textcomp}
\usepackage{enumitem}
\usepackage{float}
\usepackage{pifont}
\usepackage{soul}
\usepackage{multirow}
\usepackage{xspace}
\usepackage{footmisc}
\usepackage{setspace}
\usepackage{marvosym}
\usepackage{textcomp}
\usepackage{stfloats}
\usepackage{array}
\usepackage{bm}
\usepackage{subfigure}

\newcommand{\ie}{\emph{i.e.,}\xspace}
\newcommand{\aka}{\emph{a.k.a.,}\xspace}
\newcommand{\eg}{\emph{e.g.,}\xspace}
\newcommand{\wrt}{\emph{w.r.t.}\xspace}
\newcommand{\wo}{\emph{w/o}\xspace}

\newcommand{\ignore}[1]{}

\makeatletter
\def\@fnsymbol#1{}
\makeatother

\title{Learning to Imagine: Visually-Augmented Natural Language Generation}

\author{
	Tianyi Tang\textsuperscript{\rm{1,4}},
	Yushuo Chen\textsuperscript{\rm{1}},
	Yifan Du\textsuperscript{\rm{1}},
	Junyi Li\textsuperscript{\rm{1,3}},
	Wayne Xin Zhao\textsuperscript{\rm{1,4  \Letter}\thanks{\textsuperscript{\Letter}\ Corresponding author}\ }, \and
	Ji-Rong Wen\textsuperscript{\rm{1,2,4}} \\
	\textsuperscript{1}Gaoling School of Artificial Intelligence, Renmin University of China \\
	\textsuperscript{2}School of Information, Renmin University of China \\
	\textsuperscript{3}DIRO, Université de Montréal \\
	\textsuperscript{4}Beijing Key Laboratory of Big Data Management and Analysis Methods \\
	\texttt{steventianyitang@outlook.com \quad chenyushuo1999@foxmail.com} \\ 
	\texttt{lijunyi@ruc.edu.cn \quad \quad \ \{yifandu1999,batmanfly\}@gmail.com} \\ 
}

\begin{document}
\maketitle
\begin{abstract}
People often imagine relevant scenes to aid in the writing process. In this work, we aim to utilize visual information for composition in the same manner as humans. We propose a method, \textbf{LIVE}, that makes pre-trained language models (PLMs) \textbf{\underline{L}}earn to \textbf{\underline{I}}magine for \textbf{\underline{V}}isually-augmented natural language g\textbf{\underline{E}}neration.
First, we imagine the scene based on the text: we use a diffusion model to synthesize high-quality images conditioned on the input texts.
Second, we use CLIP to determine whether the text can evoke the imagination in a posterior way. 
Finally, our imagination is dynamic, and we conduct synthesis for each sentence rather than generate only one image for an entire paragraph. 
Technically, we propose a novel \textit{plug-and-play fusion layer} to obtain visually-augmented representations for each text. Our vision-text fusion layer is compatible with Transformer-based architecture. 
We have conducted extensive experiments on four generation tasks using BART and T5, and the automatic results and human evaluation demonstrate the effectiveness of our proposed method.
We will release the code, model, and data at the link: \url{https://github.com/RUCAIBox/LIVE}.

\end{abstract}

\section{Introduction}
Natural language generation~(NLG) is a fundamental technique for supporting a variety of downstream applications~\cite{tg_survey,llm_survey}, \eg text summarization, story generation, and data-to-text generation.  
As the mainstream NLG approach, pre-trained language models~(PLMs) can produce human-like text under the guidance of input conditions. 
Despite their success, these models are pre-trained on the text-only corpora, and they cannot well capture visually-grounded semantics, \eg visual commonsense~\cite{visual-knowledge}, making it difficult to achieve desired results when visual knowledge is required. 

To improve the generation capacity of PLMs, existing work has widely explored various methods to incorporate visual knowledge into models, which can be roughly divided into two lines of research. The first line designs specific visually-enhanced training tasks such as continual pre-training on text-image data~\cite{vlt5} or knowledge distillation with vision-language models~\cite{vlkd}.
However, these methods usually perform well only on multimodal generation tasks (\eg visual question answering) but not text generation tasks, due to the semantic disparity across modalities~\cite{voken}. As the second line, several studies retrieve or synthesize images related to the input and then fuse the image representations into PLMs~\cite{valm,inlg}.
However, they simply treat the input as a whole (even for long texts) for retrieving or synthesizing related images, which cannot sufficiently leverage fine-grained visual semantics.

Considering the above issues, we are motivated by the process of human writing where they have the ability to imagine relevant scenes from the contexts in their minds. These visual scenes convey related experiences in the world that can inspire the human's writing~\cite{imagination3,imagination2}. 
By imitating such behavior, we consider NLG as a writing process of a human, where the input text is conditioned on a set of dynamically ``\emph{imagined scenes}'', \ie synthesized images.

To this end, in this paper, we propose a novel approach, \textbf{LIVE}, that enables PLMs to \textbf{\underline{L}}earn to \textbf{\underline{I}}magine for \textbf{\underline{V}}isually-augmented natural language g\textbf{\underline{E}}neration.
Different from previous methods, our augmentation approach is relevant, selective, and dynamic.  
To be \textit{relevant}, we utilize the state-of-the-art text-to-image model, Stable Diffusion~\cite{stable-diffusion}, to synthesize realistic images for fine-grained semantic units (\ie sentences). Compared to the retrieval-based approach, our method can generate more relevant, diverse images that may not exist in real-world image databases.  
To be \textit{selective}, we evaluate the degree to which the text's meaning can be visualized in an image and only invoke the use of synthesized images when it is actually needed. 
To be \textit{dynamic}, we synthesize images for each sentence in the \emph{input} text so that the visual knowledge is more fine-grained compared to a single image for the whole input. 
In order to deeply fuse the visual knowledge of synthesized images, we propose a \textit{plug-and-play vision-text fusion layer} for Transformer-based models. 
We also design specific mechanisms to support efficient text-image cross-attention and enable the controllability of the use of visual knowledge. 

Our contributions are summarized as follows:

\textbullet~We propose a new approach, called LIVE, to learning to use synthesized images to improve natural language generation, imitating the process of human writing.

\textbullet~We propose a plug-and-play vision-text fusion layer to incorporate visual knowledge and obtain visually-augmented text representations. 

\textbullet~We verify the effectiveness of our approach with BART and T5 on four text generation tasks: LIVE consistently outperforms these PLMs, with an average improvement ratio of 2.44\%.

\begin{figure*}[!t]
	\centering
	\includegraphics[width=0.88\textwidth]{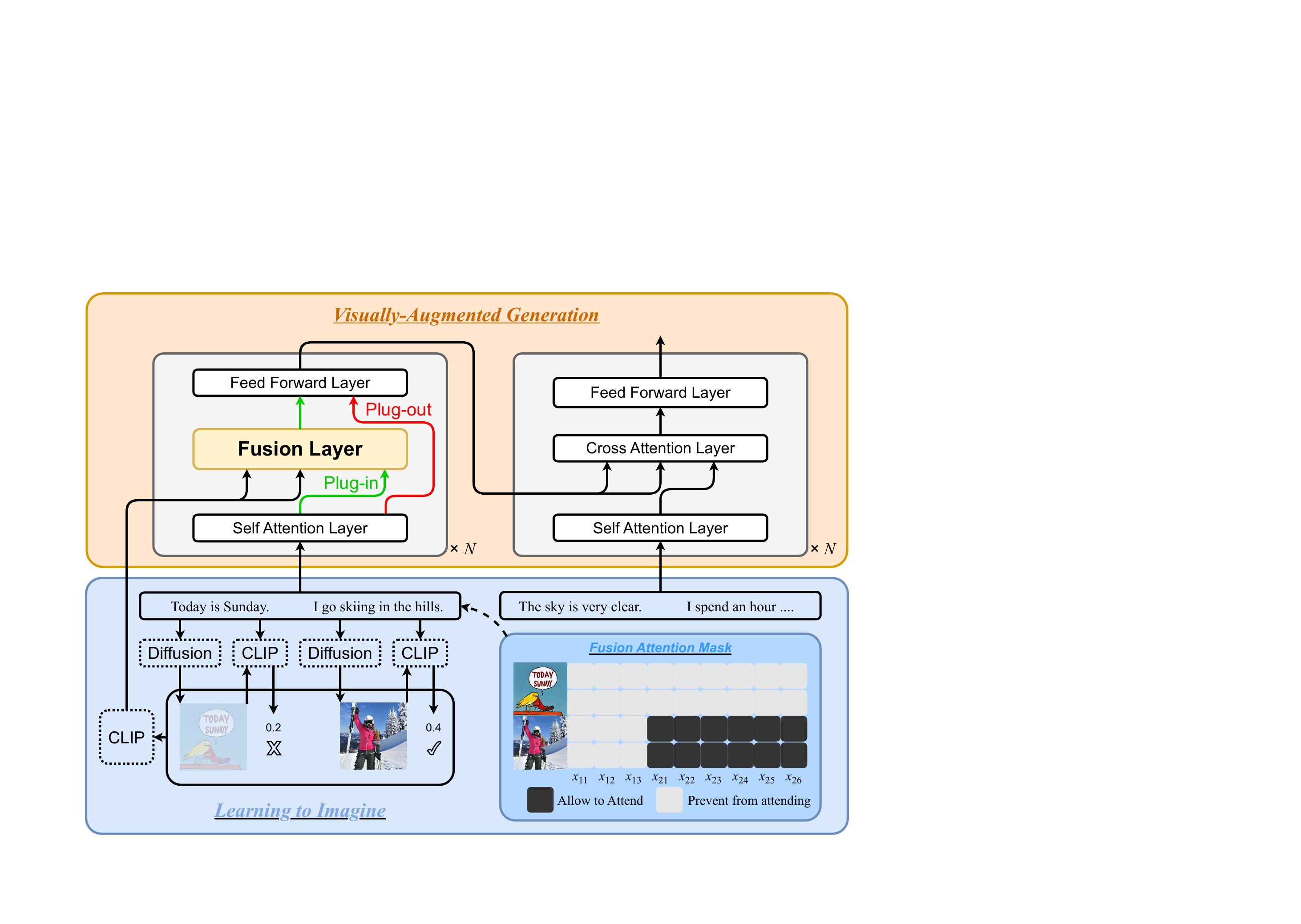}
	\caption{The overall illustration of our proposed approach \textbf{LIVE}, consisting of the text-related image generation and the plug-and-play vision-text fusion layer. The fusion attention mask means that the first sentence $x_1$ lacks visuality and will skip the fusion layer (\textcolor[RGB]{255,0,0}{red flow}), while the second sentence $x_2$ has a high visuality and each word $x_{2i}$ of $x_2$ will attend to the synthesized image to obtain visually-augmented text representations (\textcolor[RGB]{0,204,0}{green flow}).}
	\label{fig:model}
\end{figure*}
\section{Related Work}
\paragraph{Pre-Trained Models.} In recent years, large-scale pre-training has achieved remarkable success and has become the dominant technique in the NLP community~\citep{bert,t5,gpt3,llm_survey}. Pre-trained on massive text corpora, models can learn contextualized representations that include both linguistic and world knowledge~\cite{jiang-2021-know}. Since PLMs are trained with pure text corpora without connection to the visual world, vision-language pre-training (VLP) leverages image-text pairs to learn cross-modal representations~\citep{vlp_survey,vl-bert,oscar,clip}. It has been discovered that VLP models have more visual knowledge than PLMs~\cite{visual-knowledge}, however, they cannot perform well on text-only tasks such as language understanding~\cite{yun-etal-2021-vision-language}. In this work, we mainly focus on incorporating visual knowledge to enhance the performance of natural language generation tasks based on existing text-only models.

\paragraph{Visually-Augmented Language Learning.}
Considering the leakage of visual knowledge in language models, many researchers attempt to enhance text-only tasks with visual information, which is known as visually-augmented (aided or grounded) language learning. Vokenization~\cite{voken} and iACE~\cite{iace} propose to treat contextualized-related images as vokens and pre-train a text model to predict them for fusing visual information. Similarly, VidLanKD~\cite{vidlankd} extends finite image vokens to diverse video frames and employs a knowledge distillation method to acquire visual knowledge. The subsequent works leverage CLIP~\cite{clip} as the vision source to integrate visual information into PLMs via CLIP output embeddings~\cite{valm,vawi} or knowledge transfer methods~\cite{vlkd,jin-etal-2022-leveraging}. The majority of these works can outperform PLMs on language understanding tasks. As for natural language generation tasks, researchers mainly focus on finding suitable images and fusing the visual representations into text-only models using a shallow module. Some works apply generation models, such as GAN-based models~\citep{imagit,inlg} and VAE-based models~\citep{fang-feng-2022-neural}, to synthesize (latent) images, while \citet{maria}, \citet{visad}, and \citet{magic} propose to employ contextualized text embeddings to retrieve relevant images. In our work, we utilize the superior diffusion model~\cite{stable-diffusion} to synthesize high-quality images and propose a plug-and-play vision-text fusion layer to deeply integrate visual knowledge into PLMs and obtain visually-augmented text representations.

\paragraph{Multimodal Language Generation.}
Multimodal language generation aims to produce fluent and coherent text based on the input text or image. Different from unimodal language generation, the additional image serves as the background for generation. Multimodal language generation includes tasks such as image caption~\cite{coco}, visual question answering~\cite{vqa}, multimodal machine translation~\cite{multi30k}, multimodal text summarization~\cite{mm-summary}, visual dialog~\cite{vd}, and visual story telling~\cite{vist}. However, the construction of these datasets requires costly manual annotation, which hinders their widespread application. In contrast, we do not require text-image pairs as input and instead utilize Stable Diffusion~\cite{stable-diffusion}, a text-to-image model, to synthesize images for input texts.

\section{Method}

\subsection{Task Formulation}

Natural language generation (\aka text generation) aims to capture the semantic mapping relation from an input text $\mathcal{X} = \langle x_1,...,x_k,...,x_m \rangle$  to an output text $\mathcal{Y} = \langle y_1,...,y_k,...,y_n \rangle$, where $x_k$ and $y_k$ denote the $k$-th sentences of the input and output texts, respectively. In this paper, we focus on the task of \emph{visually augmented natural language generation (VA-NLG)}. Following prior works~\cite{vr,valm}, VA-NLG further assumes text-related image data can be obtained to help text generation. Here, we consider a generalized way (\eg retrieval and synthesis) to obtain the related images with an image augmenter $\mathcal{F}$, where $\mathcal{F}$ takes as input a sentence $x$ (or a text) and outputs an image $i_x$ related to $x$: $\mathcal{F}(x) \rightarrow i_{x}$.

The goal of VA-NLG is to generate readable and plausible output texts $\mathcal{Y}$ based on input texts $\mathcal{X}$ and image augmenter $\mathcal{F}$, which is formally defined as: 
\begin{equation}\label{eq-pyx}
    \text{P}(\mathcal{Y}|\mathcal{X})=\prod_{k=1}^{n} \text{P}(y_k|\mathcal{X}, y_{<k} ;\mathcal{F} ),
\end{equation}
where $y_{<k}$ denotes previously-generated sentences.

With this formulation, there are two key issues for this task: (1) how to design the image augmenter to obtain potentially useful images, and (2) how to use the augmented images for improving text generation.   
Considering the two issues, we propose \textbf{LIVE}, a general approach to augmenting NLG tasks with related images, with sentence-level image synthesis via text-to-image diffusion model (Section~\ref{sec-tig}) and  plug-and-play vision-text fusion for using augmented images (Section~\ref{sec-ppf}).

\subsection{Text-Related Image Generation}\label{sec-tig}
Although it is intuitive to augment PLMs with visual images, it is challenging to obtain appropriate and helpful images for given texts. Some previous work~\cite{vr,voken} utilizes retrieval-based methods to search images from text-image databases, such as MS COCO~\cite{coco}. However, these static image resources are limited in both \emph{quantity} and \emph{content}, which is likely to result in inaccurate image retrieval.

\paragraph{Synthesizing Relevant Images.}
To circumvent the limitation of static image resources, we instead propose to automatically generate images for given texts by leveraging text-to-image generation models. In contrast to previous works that utilize GAN-based~\cite{vqgan} or auto-regressive~\cite{ofa} generation models, we use the state-of-the-art Stable Diffusion model~\cite{stable-diffusion}, a probabilistic diffusion model guided by CLIP-encoded input text representations, to synthesize high-quality images. 
With Stable Diffusion, we can flexibly perform image generation based on different text units. Here, we consider \emph{sentences} as synthesis units, which contain a moderate amount of information for an image. Compared with the previous work that synthesize a single image for the whole input, our sentence-level generation is more fine-grained. It is inspired by the writing behavior of people: one would switch the imagined scenes for different sentences.

For each input sentence $x_k$, we apply Stable Diffusion to synthesize its corresponding creative image $i_{x_k}$.
Equipped with the acceleration method of DDIM~\cite{ddim}, Stable Diffusion is able to synthesize photographic images normally in  50 steps~\cite{stable-diffusion}. 
In practice, we empirically find that using a 25-step synthesis can usually lead to a decent performance in our task (see Section~\ref{exp-image} for more analysis about the diffusion quality and efficiency).

\paragraph{Evaluating the Text Visuality.} 
Although the generation-based method is flexible to produce images on various topics, not all texts can inspire the generative model to generate meaningful images, such as the rule text \textit{``ACL 2023 requires all papers to have a clear discussion of limitations''}. Only texts with visually rich content can be associated with images. Previous work usually synthesizes or retrieves images without considering the visuality of texts, tending to incorporate irrelevant or noisy images. 
However, it is difficult to directly measure the visuality of a text. As a compromise, we estimate the similarity score in a posterior way between a sentence $x_k$ and  a synthesized image $i_{x_k}$ using CLIP~\cite{clip}:
\begin{equation} \label{eq-sim}
    \gamma = \text{CLIP}(x_k,  i_{x_k}) \in \left[ -1,1 \right].
\end{equation}
CLIP is a vision-language model pre-trained on a massive amount of text-image pairs using contrastive learning which excels at evaluating the similarity between text and image. 
In our work, we manually set a threshold value $\theta$. If $\gamma$ exceeds the threshold value, the text is considered to have high visuality; otherwise, we consider that the text has weak visuality and discard the synthesized image. We will discuss the influence of $\theta$ in Section~\ref{exp-theta}.

\subsection{Plug-and-Play Vision-Text Fusion}\label{sec-ppf}

After synthesizing relevant images for given texts, we study how to leverage  
visual images  for improving text generation. Instead of using VLP models, we 
aim to fuse the visual knowledge into a PLM-based backbone,   since text generation is essentially a language modeling task. 
To enhance the cross-modality fusion, we propose  a plug-and-play vision-text fusion module to obtain deeply-fused visually-augmented text representations.

\ignore{the next step is to fuse the visual context with texts for improving text generation.
A direct idea is to employ VLP models, which were pre-trained on paired image-text data. 
However, the language-only data for text generation has a significantly different distribution from the paired image-text data~\cite{voken}.
Therefore, we explore to integrate the visual information into existing PLMs, which can not only preserve the language generation capacity of PLMs but also benefit from the visual knowledge. 
Rather than the shallow vision-text fusion in existing work~\cite{vr,valm}, we propose a novel plug-and-play vision-text fusion module to obtain deeply-fused visually-augmented text representations. The fusion module is a controllable and efficient layer that works well with existing PLMs.
}

\paragraph{Vision-Text Fusion for PLMs.} Our fusion module is a plug-and-play attention layer for Transformer-based~\cite{transformer} models, such as BART~\cite{bart} and T5~\cite{t5}. We insert the fusion layer after the self-attention layer in the encoder. Our fusion layer is a layer-wise cross-attention module to augment the word representations with visual information.  
In particular, for a sentence $x_k$ and the corresponding synthesized image $i_{x_k}$, we first utilize CLIP to encode the image into patch representations $\mathbf{I}_k \in \mathbb{R}^{p \times d}$. Then, we feed the sentence into the Transformer model and obtain the output representation $\mathbf{S}_{k,l}$ for the self-attention sub-layer in the $l$-th layer of the encoder.
Finally, we pass $\mathbf{S}_{k,l}$ to our $l$-th plug-and-play fusion layer to obtain the visually-augmented text representations:
\begin{equation}
    \mathbf{F}_{k,l}=
    \begin{cases}
        \text{FusionLayer}_l(\mathbf{S}_{k,l}, \mathbf{I}_k, \mathbf{I}_k), & \gamma \geq \theta \\
        \mathbf{S}_{k,l}, & \gamma < \theta
    \end{cases},
\end{equation}
where $\gamma$ is the similarity score computed in Equation~\ref{eq-sim}, and $\text{FusionLayer}_l$ conducts multi-head attention on the query, key, and value matrices, followed by residual connection and layer normalization. Here, we introduce $\gamma$ to control whether a generated image will be used or not. 

In general, such a fusion layer can be applied to various Transformer-based PLMs and LLMs. 
Note that each sentence attends to no more than one image, as depicted in the attention matrix in Figure~\ref{fig:model}. Compared to simply concatenating images and text as input~\cite{maria}, our cross-attention-based mechanism is more efficient while maintaining performance (see Section~\ref{exp-ab}). 
Besides, our fusion is more controllable and can achieve fine-grained cross-attention. For example, we can choose only nouns to be attended with images since they contain more visual information (see Section~\ref{exp-ab}).

\subsection{Optimization}

In order to achieve decent performance, we can pre-train the key component of our approach, \ie the fusion layer (Section~\ref{sec-ppf}), with text-image paired datasets. Specially, we collect the image caption datasets MS COCO~\cite{coco}, Flickr30k~\cite{flickr}, CC3m~\cite{cc3m}, and Visual Genome~\cite{vg} as text-image pairs, and utilize the caption text to synthesize images using Stable Diffusion to enrich the pre-training pairs. In this way, we can obtain 9 million text-image pairs in total. Then, we apply image-based denoising autoencoding as the pre-training objective, which teaches the model to recover the caption based on a noisy text. Such a pre-training strategy can make the fusion layer better map the visual knowledge into text space.

Next, we describe the overall optimization process of our approach. During pre-training, we freeze the PLM backbone and only pre-train the fusion layer; therefore, if we plug-out the fusion layer, the PLM retains its original language generation ability. The fusion layer is a lightweight module and has 18M parameters for BART\textsubscript{\textsc{base}} (140M).
During fine-tuning, we utilize Stable Diffusion and CLIP models to synthesize images and compute similarity scores. These operations can be done offline for efficiency, and the diffusion and CLIP models will not be updated. We only need to fine-tune the whole PLM as usual, in addition to the small pre-trained fusion layer.

\section{Experiment}

\subsection{Experimental Setup}

\subsubsection{Dataset}
We conduct experiments on four text generation datasets with diverse tasks and domains:

\textbullet~E2E~\cite{e2e} is a data-to-generation task with the aim of converting multiple input meaning representations into fluent texts.

\textbullet~CommonGen~\cite{cg} requires the model to generate a coherent and reasonable text given a collection of common concepts.

\textbullet~SAMSum~\cite{samsum} is a dialogue summarization dataset that evaluates the model's summary and dialogue understanding abilities.

\textbullet~ROCStories~\cite{roc} consists of five-sentence stories, and we utilize the first sentence as input to generate the remaining four.

The details of the statistics and license of each dataset are listed in Table~\ref{tab:data}. For each dataset, we utilize NLTK\footnote{\url{https://www.nltk.org/}} to tokenize the input texts into sentences, except that we treat each key-value pair in the input as a sentence for the E2E dataset.

\begin{table}[t]
	\centering
	\resizebox{1\columnwidth}{!}{
		\begin{tabular}{rrrrl}
			\toprule
			\textbf{Dataset} &
			\multicolumn{1}{c}{\textbf{\#Train}} &
			\multicolumn{1}{c}{\textbf{\#Valid}} &
			\multicolumn{1}{c}{\textbf{\#Test}} &
			\multicolumn{1}{c}{\textbf{License}}\\
			\midrule
			CommonGen      & 67,389  & 993    & --     & MIT  \\
			E2E            & 42.061  & 547  & 630     & CC BY-SA 4.0 \\
			ROCStories     & 176,688 & 9,816  & 4,909  & N/A \\
			SAMSum         & 14,732  & 818    & 819    & CC BY-NC-ND 4.0 \\
			\bottomrule
	\end{tabular}}
	\caption{The statistics and licenses of datasets.}
	\label{tab:data}
\end{table}

\subsubsection{Evaluation Metrics}
We adopt five automatic metrics, namely BLEU~\cite{bleu}, ROUGE~\cite{rouge}, CIDEr~\cite{cider}, SPICE~\cite{spice}, and Distinct~\cite{distinct}, to compare the performance of different methods. BLEU, ROUGE, and CIDEr compute the n-gram overlap between the candidate text and the reference text(s). SPICE further takes semantic meaning into consideration. Distinct mainly evaluates the diversity of the generated texts and is always used in open-ended generation tasks, such as story generation. We also conduct the human evaluation in Section~\ref{exp-human}.

\subsubsection{Baseline Models}
We utilize two commonly used text generation PLMs, BART~\cite{bart} and T5~\cite{t5}, as text-only baselines. We further compare them to two multimodal VLP models:

\textbullet~BLIP~\cite{blip} uses a multimodal mixture of encoder-decoder with the objectives of text-image contrast, text-image matching, and language modeling on bootstrapped text-image pairs.

\textbullet~OFA~\cite{ofa} unifies text and image modalities using a unified architecture and multi-task sequence-to-sequence learning. In addition, we consider a variant and attempt to use OFA with only text, denoted by OFA \wo image.

We integrate our LIVE framework with BART and T5, and consider the following visually-augmented methods as comparisons:

\textbullet~VL~\cite{vlt5} adds visual embeddings for the original BART and T5 and conducts continued pre-training on text-image pairs.

\textbullet~iNLG~\cite{inlg} guides the PLM with the machine-generated image as the visual prefix. Since iNLG does not offer a T5 version, we can only combine it with BART for comparison.

\begin{table*}[!th]
	\small
	\centering
	\begin{tabular}{lccccccccccc}
		\toprule
		\multirow{2.5}{*}{Methods} & \multicolumn{3}{c}{\textbf{E2E}} & \multicolumn{3}{c}{\textbf{CommonGen}} & \multicolumn{3}{c}{\textbf{SAMSum}} & \multicolumn{2}{c}{\textbf{ROCStories}} \\ 
		\cmidrule(lr){2-4} \cmidrule(lr){5-7} \cmidrule(lr){8-10} \cmidrule(lr){11-12} 
		& B-4 & R-L & ME & B-4 & CIDEr & SPICE & R-1 & R-2 & R-L & B-1 & D-4 \\ 
		\midrule
		\textbf{BLIP}          & 45.05 & 54.35 & 34.84 & 13.30 & 5.84 & 18.62 & 22.54 & 4.07 & 20.56 & 28.29 & 66.93 \\
		\textbf{OFA}           & 67.20 & 69.18 & 45.12 & 29.34 & 15.48 & 30.79 & 47.42 & 23.20 & 43.45 & 31.70 & 68.16 \\
		\textbf{OFA} \wo image & 67.63 & 69.08 & 45.19 & 29.54 & 15.46 & 30.84 & 48.12 & 23.33 & 43.81 & 32.51 & 70.99 \\
		
		\midrule[0.3pt]
		
		\textbf{BART}          & 67.38 & 69.57 & 45.04 & 30.30 & 16.05 & 31.16 & 49.92 & 25.55 & 45.61 & 32.98 & 76.77 \\
		\textbf{VL-BART}       & 68.53 & 69.57 & 45.17 & 29.51 & 15.19 & 29.54 & 45.02 & 20.22 & 40.83 & 32.76 & 76.32 \\
		\textbf{iNLG-BART}     & 64.71 & 67.19 & 43.14 & 29.80 & 15.80 & 30.78 & 50.75 & 26.20 & 46.36 & 33.25 & 50.87 \\
		\textbf{LIVE-BART}     & \textbf{69.24} & \textbf{70.59} & \textbf{45.60} & \textbf{31.47} & \textbf{16.55} & \textbf{31.89} & \textbf{51.31} & \textbf{26.67} & \textbf{47.08} & \textbf{33.46} & \textbf{79.98} \\
		
		\midrule[0.3pt]
		
		\textbf{T5}            & 66.54 & 68.02 & 44.71 & 26.70 & 15.66 & 30.96 & 49.27 & \textbf{25.30} & 45.18 & 33.14 & 75.11 \\
		\textbf{VL-T5}         & 66.96 & 70.09 & 44.66 & 27.29 & 15.31 & 29.78 & 49.91 & 24.95 & 45.20 & 33.07 & 75.09 \\
		\textbf{LIVE-T5}       & \textbf{68.34} & \textbf{71.11} & \textbf{46.09} & \textbf{27.94} & \textbf{15.84} & \textbf{31.36} & \textbf{49.99} & 25.16 & \textbf{45.84} & \textbf{33.22} & \textbf{77.28} \\
		\bottomrule
	\end{tabular}
	
	\caption{The results of four text generation tasks. B, R, ME, and D are short for BLEU, ROUGE, METEOR, and Distinct, respectively. The best results are highlighted in \textbf{bold}. These setups and abbreviations are the same below.}
	\label{tab:main}
\end{table*}

\subsubsection{Implementation Details} \label{exp-detail}
For all baselines, we utilize the base versions of PLMs, \ie BART\textsubscript{\textsc{base}}, T5\textsubscript{\textsc{base}}, BLIP\textsubscript{\textsc{base}}, and OFA\textsubscript{\textsc{base}}, which have a comparable number of parameters to ensure a fair comparison. For BLIP, OFA, VL-BART, and VL-T5, we provide the same synthesized image as our method, and we fine-tune them similarly to how they perform VQA tasks. For iNLG, we utilize its official implementation\footnote{\url{https://github.com/VegB/iNLG}}.

As for our method, we employ Stable Diffusion v1.4 with half precision\footnote{\url{https://huggingface.co/CompVis/stable-diffusion-v1-4}} to synthesize images in 25 timesteps for efficiency. Then, we adopt CLIP-ViT-B/32 to judge the similarity between text-image pairs and extract image features. We empirically set the threshold value $\theta=0.27$. After extraction, an MLP layer is appended to project the image representation into the text space and obtain an image representation $\bm{I}_i \in \mathbb{R}^{50 \times 768}$. The aforementioned operations can be performed offline for efficiency. 

In the pre-training stage of our fusion layer, we mask 50\% of the input text with span lengths drawn from a Poisson distribution with $\lambda=3.5$ for BART and force the model to recover the input with the image. As for T5, we split the caption into two parts and train the model to generate the second part using the first part and the image. We pre-train the fusion layer with a batch size of 384, optimize BART using AdamW~\cite{adamw} with a constant learning rate of $1 \times 10^{-5}$, and optimize T5 using Adafactor~\cite{adafactor} with a learning rate of $1 \times 10^{-3}$.

In the fine-tuning stage, we tune the entire model, including the PLM backbone and the fusion layer. We set the batch size to 32 and employ the same optimizer and learning rate as in pre-training. We optimize the model using cross-entropy sequence-to-sequence loss with a label smoothing factor~\cite{label} of 0.1. During inference, we choose the checkpoint with the highest validation metric score for generation. During generation, we apply beam search with a beam size of 5 for E2E, CommonGen, and SAMSum, while utilizing the nucleus sampling with $p=0.9$ and $t=0.7$ for ROCStories. 

All the experiments are conducted using the text generation library TextBox~\cite{textbox} on NVIDIA GeForce RTX 3090 24GB GPUs using Ubuntu 20.04.1 SMP. \textit{All these hyper-parameters are identical for our method and baselines.}

\subsection{Experimental Results} \label{exp-main}

Based on the results in Table~\ref{tab:main}, we can find that:

Firstly, the results of multimodal models (\ie BLIP and OFA) cannot achieve satisfactory results when compared with text-only models (\ie BART and T5) on pure text tasks. This finding further proves the existence of semantic disparity~\cite{voken} across modalities of generation tasks. OFA without images even outperforms OFA with images slightly, which indicates that images may be a burden for text generation tasks when the fusion method is not appropriate.

Secondly, the visually-augmented methods (\ie VL-BART, VL-T5, and iNLG) can achieve superior performance than their base PLMs on certain tasks but cannot achieve overall improvement on all tasks. A major reason might be that they synthesize only one image for each input without considering its relevance and sentence-level semantics.

Finally, our LIVE method can outperform all baselines on all four text generation tasks. Equipping BART with our LIVE method, LIVE-BART can outperform its text-only counterpart BART by 2.80\% in ratio. LIVE can also work with T5, yielding an average improvement of 2.08\%. These automatic results demonstrate the effectiveness and compatibility of our text-related image generation approach and plug-and-play fusion layer.

\section{Further Analysis}
In this section, we conduct various experiments to test the efficacy of our methods. The tuning details are identical to those introduced in Section~\ref{exp-detail}.

\begin{table*}[!t]
    \small
    \centering
    \begin{tabular}{lcccccccccccc}
        \toprule
        \multirow{2.5}{*}{Methods} & \multicolumn{3}{c}{\textbf{0.1\%}} & \multicolumn{3}{c}{\textbf{0.3\%}} & \multicolumn{3}{c}{\textbf{1\%}} & \multicolumn{3}{c}{\textbf{3\%}} \\ 
        \cmidrule(lr){2-4} \cmidrule(lr){5-7} \cmidrule(lr){8-10} \cmidrule(lr){11-13} 
        & B-4 & R-L & ME & B-4 & R-L & ME & B-4 & R-L & ME & B-4 & R-L & ME \\ 
        \midrule
        \textbf{BART}      & 50.58 & 57.95 & 32.37 & 56.18 & 62.34 & 36.02 & 62.11 & 66.38 & 39.34 & 65.25 & 68.15 & 42.18 \\
        \textbf{iNLG-BART} & 28.40 & 53.89 & 25.98 & 39.15 & 58.63 & 30.05 & 48.66 & 62.12 & 33.31 & 61.74 & 65.75 & 38.05 \\
        \textbf{LIVE-BART} & \textbf{51.67} & \textbf{60.41} & \textbf{33.06} & \textbf{60.87} & \textbf{64.32} & \textbf{38.22} & \textbf{63.31} & \textbf{67.00} & \textbf{40.30} & \textbf{65.99} & \textbf{69.08} & \textbf{43.00} \\
        \bottomrule
    \end{tabular}
    
    \caption{The few-shot experiments on the E2E dataset.}
    \label{tab:few-shot}
\end{table*}

\subsection{Few-Shot Results} \label{exp-few}
We investigate the performance of our LIVE methods in a low-resource situation. We keep 0.1\%, 0.3\%, 1\%, and 3\% of the training set for the E2E dataset. For each split, we choose five independent groups to decrease the randomness. From the results in Table~\ref{tab:few-shot}, we can observe that our methods remarkably boost the performance under few-shot settings compared with baselines, especially in extreme situations (0.1\% and 0.3\%). We assume that synthesized images can provide visual knowledge as a supplement when training data is scarce.

\subsection{Ablation Study} \label{exp-ab}
To examine the effectiveness of the different factors of our LIVE methods, we conduct four groups of experiments for ablation. The results are reported in Tables~\ref{tab:ablation} and~\ref{tab:ablation2}. First, we can see that the \textit{pre-training} of the vision-text fusion layer is beneficial. 

Second, we replace the \textit{image augmenter} $\mathcal{F}$ Stable Diffusion with two variants: a text-image retriever CLIP~\cite{clip} and a text-to-image synthesizer VQGAN~\cite{vqgan}. We can find that the synthesis-based methods are superior to the retrieval-based ones since they can generate relevant images which may not exist in a static database. Compared with VQGAN, Stable Diffusion can synthesize high-quality images and provide more visual knowledge for text generation. 

\begin{table}[!t]
	\small
	\centering
	\begin{tabular}{lccc}
		\toprule
		& B-4   & R-L   & ME \\ 
		\midrule
		\textbf{LIVE-BART}     & 69.24 & 70.59 & 45.60 \\
		\quad \wo pre-training & 68.02 & 69.72 & 45.33\\
		\midrule
		\multicolumn{4}{l}{\textit{Image augmenter}} \\
		\quad CLIP             & 65.70 & 68.65 & 44.63 \\
		\quad VQGAN            & 67.13 & 69.42 & 45.15 \\
		\midrule
		\multicolumn{4}{l}{\textit{Fusion method}} \\
		\quad Concatenation    & 67.30 & 69.37 & 45.12 \\
		\quad Self-attention & 68.08 & 69.72 & 45.28 \\
		\bottomrule
	\end{tabular}
	
	\caption{Ablation analysis on the E2E dataset. The experiments with different image augmenters and fusion methods are conducted without pre-training.}
	\label{tab:ablation}
\end{table}

\begin{table}[!t]
	\small
	\centering
	\begin{tabular}{lccc}
		\toprule
		\textbf{Image source}         & B-4   & R-L   & ME    \\ 
            \midrule
            Sent-level (Ours) & 69.24 & 70.59 & 45.60 \\
		\midrule[0.3pt]
		Doc-level              & 68.25 & 70.24 & 45.26 \\
		Selective sent-level   & 69.30 & 70.62 & 45.69 \\
		Word-level             & 67.67 & 69.58 & 45.36 \\
		\bottomrule
	\end{tabular}
	
	\caption{Further analysis on the different granularities of different image synthesis strategies.}
	\label{tab:ablation2}
\end{table}

Third, we investigate the \textit{fusion method} of visual representations and make two variants of our cross-attention-based fusion. ``Concatenation'' means to concatenate the image representations and the encoder output as the input for the decoder, while ``Self-attention'' means to concatenate the image representations and the text representations as the input for the encoder. The results indicate that the deep fusion of text and vision representations is beneficial and the cross-attention-based method and self-attention-based method are comparable, which is consistent with~\citet{vlp_survey}. Thus, we utilize cross-attention as the fusion method because it is more efficient and controllable.

Finally, we explore our dynamic and controllable fusion layer. To be dynamic, we synthesize one image for each sentence in the input (denoted as ``Sent-level'') and attempt two variants that synthesize one image for the whole document (``Doc-level'') or each word in the document (``Word-level''). The results prove the effectiveness of our sentence-level synthesis compared with previous method~\cite{inlg} that only generates one image for the input. However, too many images actually lead to poor performance. In addition, we investigate a fine-grained cross-attention based on sentence-level synthesis (``Selective sent-level''). We only make noun words visually-augmented and make the other words skip the fusion layer. The results show that the fine-grained fusion may be promising, and we leave it for future work.

\begin{figure}[!t]
	\centering
	\includegraphics[width=0.75\columnwidth]{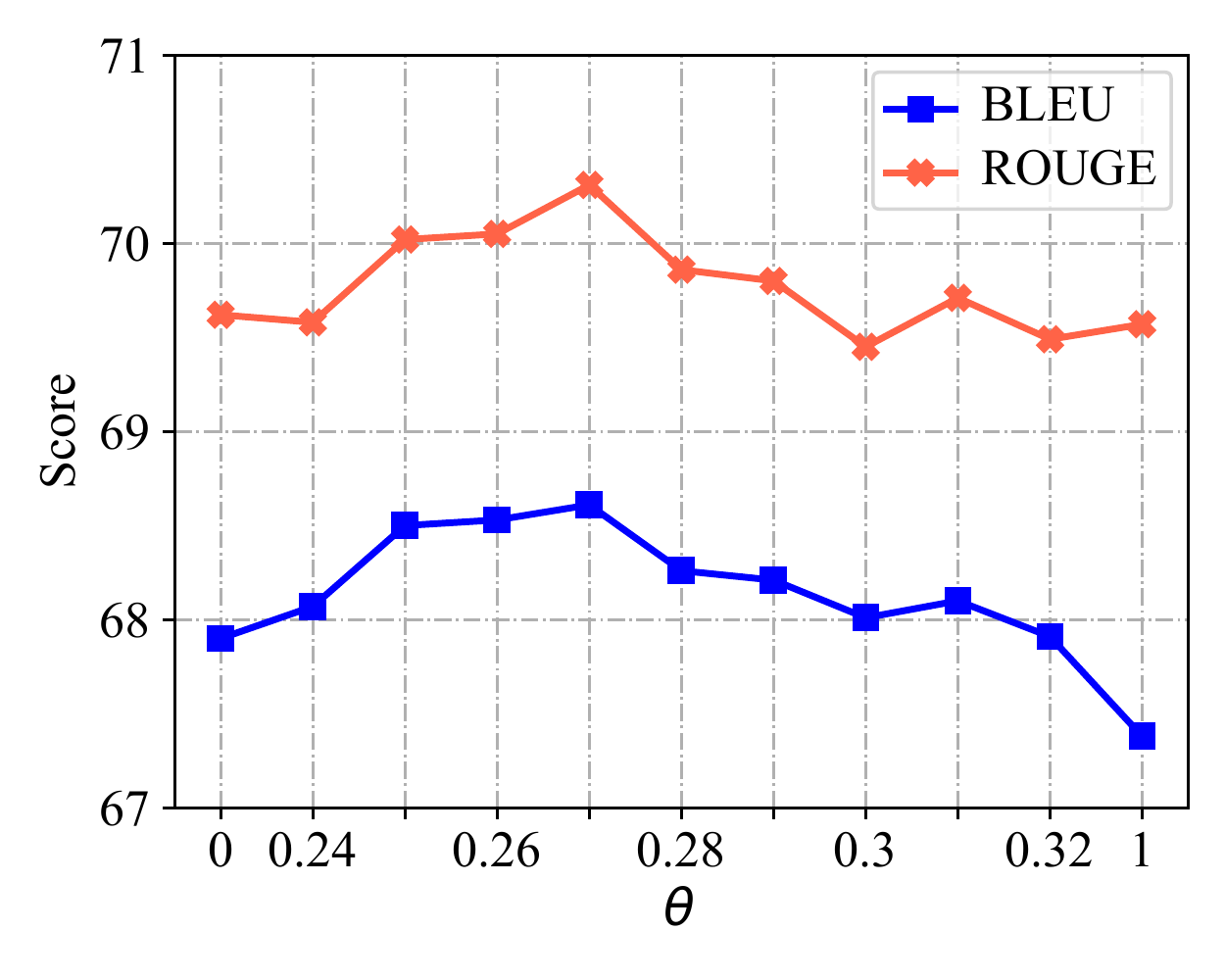}
	\caption{Varying the similarity threshold value $\theta$.}
	\label{fig:theta}
\end{figure}

\begin{figure}[!t]
	\centering
	\includegraphics[width=0.75\columnwidth]{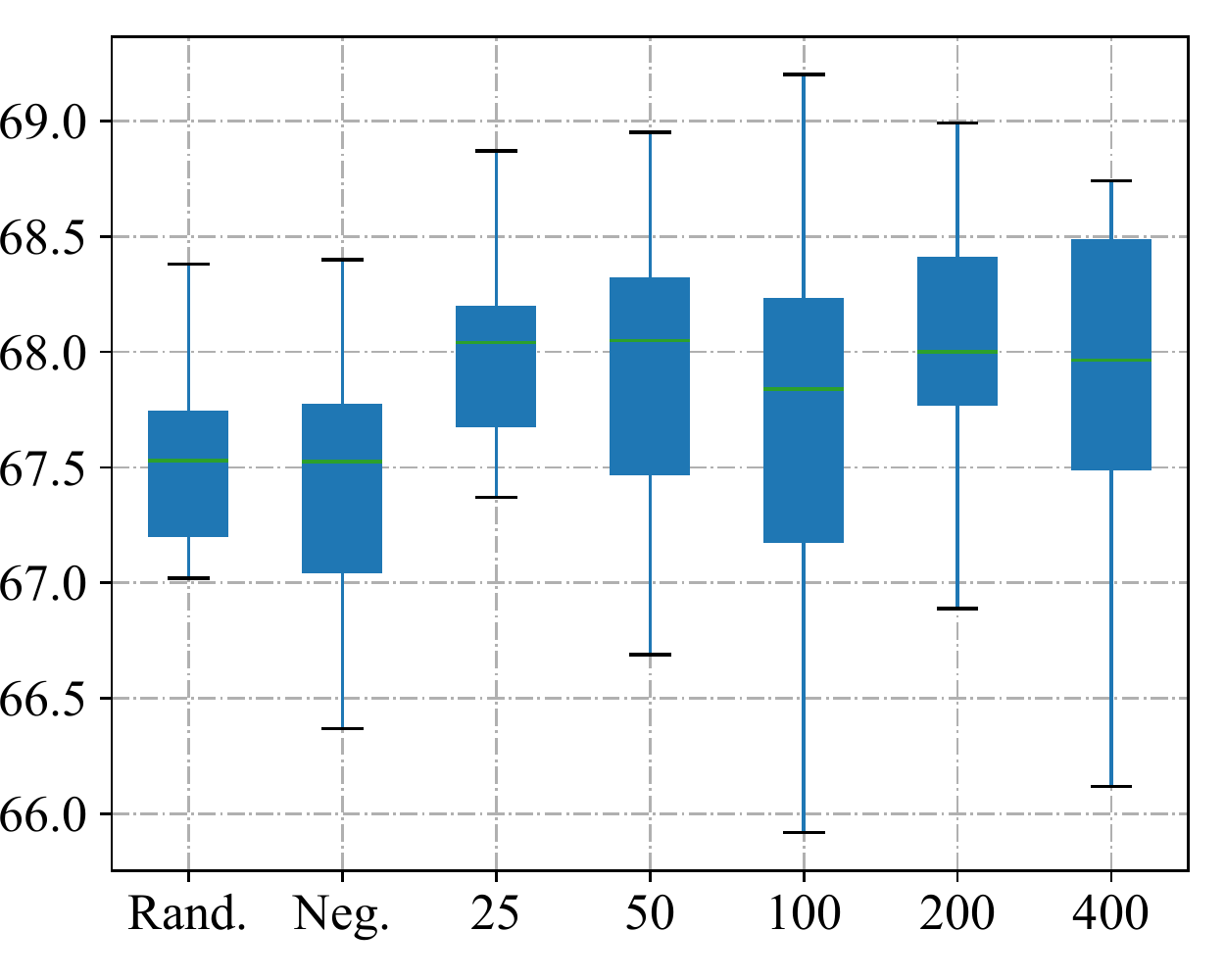}
	\caption{Varying the number of diffusion steps.}
	\label{fig:image}
\end{figure}

\subsection{Model Sensitivity \wrt the Similarity Threshold Value $\theta$} \label{exp-theta}
In Section~\ref{sec-tig}, we set a threshold value $\theta$ to measure the text visuality. Here, we investigate the model's performance when $\theta$ varies. If $\theta=0$, all the sentences will be visually-augmented. If $\theta=1$, all the sentences will not be visually-augmented, and it degenerates to text-only BART. As shown in Figure~\ref{fig:theta}, LIVE-BART with $\theta=0.27$ achieves the best performance, and we find that $0.27$ is close to the median of text visuality scores, \ie nearly half of the sentences will be augmented and the others will not be. Therefore, we set $\theta=0.27$ for our LIVE methods in experiments.

\subsection{Model Sensitivity \wrt the Synthesized Images} \label{exp-image}
In this subsection, we first demonstrate that visual information is truly favorable for text generation. Following the previous works~\cite{vr}, we replace the image representations with random noise or utilize the input text as a negative prompt to synthesize irrelevant images. The results in Figure~\ref{fig:image} further prove the necessity of visual knowledge for text generation. Moreover, we vary the number of diffusion steps since it is a trade-off between synthesis quality and efficiency. Surprisingly, increasing the diffusion steps will not lead to performance gains. We speculate that diffusion with certain steps can provide enough visual knowledge for the PLM, and more steps may just help to achieve higher resolution. Thus, we only synthesize for 25 steps considering the efficiency.

\begin{table}[!t]
	\small
	\centering
	\resizebox{1\columnwidth}{!}{
	\begin{tabular}{rcccc}
		\toprule
		\textbf{Datasets} & \textbf{LIVE+BART wins} & \textbf{Ties} & \textbf{BART wins} \\ 
		\midrule
		E2E               & 29\%                  & 56\%              & 15\%                   \\
		CommonGen         & 24\%                  & 58\%              & 18\%                   \\
		SAMSum            & 40\%                  & 34\%              & 26\%                   \\
		ROCStories        & 48\%                  & 11\%              & 41\%                   \\
		\bottomrule
	\end{tabular}}
	\caption{Human evaluation on four generation tasks.}
	\label{tab:human}
\end{table}

\subsection{Human Evaluation} \label{exp-human}
Considering that the automatic evaluation may be inconsistent with human judgments, we further invite five college students to assess the generated texts. We randomly choose 100 samples from the test set of each dataset and showcase the generated texts of both BART and LIVE-BART. The annotators should choose which one is better or choose a tie based on their subjective feelings. From the results in Table~\ref{tab:human}, we can observe that our LIVE method can make BART generate more satisfactory texts in all tasks.

\section{Conclusion}
In this paper, we present the \textbf{LIVE} method for natural language generation. First, we propose an imagination-based method, imitating the process of human writing. It is a relevant, selective, and dynamic approach that leverages Stable Diffusion to synthesize images for each input sentence and discard the images with lower text visuality computed by CLIP. Furthermore, we introduce a plug-and-play vision-text fusion layer to deeply incorporate visual knowledge into PLMs and obtain visually-augmented text representations for text generation. Extensive experiments have demonstrated that our LIVE methods are compatible with two PLMs (\ie BART and T5) and can achieve superior performance over all the baseline models.

In future work, we will investigate how to synthesize more relevant images based on the input prompt and design a finer fusion method for better aligning different words and images. We will also attempt to extend our methods to more tasks (\eg language understanding) and PLMs (\eg BERT). Besides, it is meaningful to explore the probability of combining our LIVE method with existing large language models~\cite{llm_survey} to enhance their representation and generation capabilities. 

\section*{Acknowledgment}
This work was partially supported by National Natural Science Foundation of China under Grant No. 62222215, Beijing Natural Science Foundation under Grant No. 4222027, and Beijing Outstanding Young Scientist Program under Grant No. BJJWZYJH012019100020098. Xin Zhao is the corresponding author.

\section*{Limitations}
We only conduct experiments on four natural language generation tasks without considering the expandability to more NLP tasks, such as language understanding or reasoning. It is also meaningful to investigate the robustness of our methods with different text formats (\eg text length and literary form), \ie examine which situations and why our methods can achieve better performance. Due to the limitation of computing power, we do not explore the effectiveness of our methods under different PLMs with various scales. Besides, we utilize CLIP to evaluate the text visuality and encode images into representations, and this is also interesting to research which vision encoder has higher suitability with PLMs.

\bibliography{ref}
\bibliographystyle{acl_natbib}

\end{document}